# A Variational Approximation for Bayesian Networks with Discrete and Continuous Latent Variables

**Kevin P. Murphy**
Computer Science Division, Univ. of California, Berkeley, CA 94720
murphyk@cs.berkeley.edu

## Abstract

We show how to use a variational approximation to the logistic function to perform approximate inference in Bayesian networks containing discrete nodes with continuous parents. Essentially, we convert the logistic function to a Gaussian, which facilitates exact inference, and then iteratively adjust the variational parameters to improve the quality of the approximation. We demonstrate experimentally that this approximation is much faster than sampling, but comparable in accuracy. We also introduce a simple new technique for handling evidence, which allows us to handle arbitrary distributions on observed nodes, as well as achieving a significant speedup in networks with discrete variables of large cardinality.

## 1 Introduction

Many probabilistic models naturally contain discrete and continuous variables. (Such models are sometimes called "hybrid".) Unfortunately, exact inference is only possible when all the continuous variables are Gaussian and have no discrete children. If we want to allow discrete children of continuous parents (e.g., to model threshold phenomena), the standard approach is to discretize all the variables [FG96, KK97] or resort to sampling [SP90, GRS96]. The problem with discretization is that, to get good accuracy, we must quantize finely, which makes inference slow; this problem is especially acute in high-dimensional state spaces. The problem with sampling is similar: to get good accuracy, we must take many samples, which is slow. In this paper, we introduce a variational approximation to handle the case of discrete children of continuous parents, which is faster and more accurate, since all the distributions that can be handled exactly are handled exactly. We also introduce a new approach to dealing with evidence, which allows us to handle arbitrary distributions on observed nodes.

We present our results in the context of the junction tree algorithm, which is widely considered to be the most efficient and most general inference algorithm for graphical models [SAS94]. In particular, it allows us to compute the marginals on all $N$ families — a prerequisite for efficient parameter and structure learning — in two passes over the graph, whereas other, query-driven (goal-directed) algorithms, such as bucket-elimination [Dec98] and SPI [CF91, CF95], would take $N$ passes. In addition, the junction tree algorithm allows us to handle graphs with undirected cycles, unlike some previous work on networks with continuous variables [DM95, AA96] which was restricted to polytrees.

The structure of the paper is as follows. We start by describing some popular conditional probability distributions (CPDs) for nodes in hybrid networks. In Section 3, we give a brief overview of the junction tree algorithm, and in Sections 4 through 6, we review aspects of it that are specific to hybrid networks. In Section 7, we explain our variational approximation, in Section 8 we introduce our new approach to handling evidence, in Section 9 we discuss the computational complexity of inference in hybrid BNs, and in Section 10, we present some experimental results to assess the quality of our approximation. We finish by discussing future work.

## 2 CPDs for hybrid networks

For any directed graphical model, we must define the conditional distribution of each node given its parents: see Table 1 for some examples.

For discrete nodes with discrete parents, the simplest representation is a table (called a Conditional Probability Table, or CPT), which defines $\Pr(R = i | Q = j) \stackrel{\text{def}}{=} \theta_{ij}$. (A note on nomenclature: we will use $Q$ to represent a discrete parent, $R$ to represent a discrete child, $X$ to represent a continuous parent, and $Y$ to represent a continuous child.) If there are multiple parents, $Q_1, \ldots, Q_p$, we can use a multi-dimensional table, although this requires specifying $O(2^p)$ parameters (assuming for simplicity that each discrete node is binary). There are other representations which require fewer parameters (e.g., noisy-OR, neural networks), and hence are easier to learn, but we don't discuss them here.

Now let us consider the case of continuous nodes with continuous parents. (Without loss of generality, we can assume the child has only one continuous parent, since if it has more than one, we can aggregate them into a single vector-valued node.) The simplest such example is a Gaussian whose mean is a linear function of its parent's value:

$$P(Y = y | X = x) = N(y; \mu + Wx, \Sigma)$$

| Child/Parent | Discrete | Continuous |
|---|---|---|
| Discrete | Tabular, noisy-OR, decision tree | Probit, logistic, softmax |
| Continuous | Conditional Gaussian | Linear Gaussian |

Table 1: Some popular conditional probability distributions. If a node has both discrete and continuous parents, we can create a mixture distribution.

where $W$ is the weight or regression matrix,

$$N(y; \mu, \Sigma) = C(\Sigma) \exp\left[-\tfrac{1}{2}(y-\mu)'\Sigma^{-1}(y-\mu)\right]$$

is the Normal (Gaussian) distribution, and

$$C(\Sigma) = (2\pi)^{-n/2}|\Sigma|^{-\tfrac{1}{2}}$$

is the normalizing constant ($n$ is the number of rows/columns in $\Sigma$), which ensures $\int_y N(y; \mu, \Sigma) = 1$.

Networks in which all the variables have this kind of linear Gaussian distribution were studied in [SK89]. If the continuous child (also) has discrete parents, we can specify a Gaussian for each value of the discrete parents; this is called a Conditional Gaussian (CG) distribution. Note that a CG distribution can be used to approximate arbitrary continuous distributions.

Finally, we consider the case of discrete nodes with continuous parents. There are two popular models for the conditional distribution of a discrete binary variable $R \in \{0, 1\}$ given a continuous (vector-valued) parent $X$, called logistic and probit, which are defined as follows:

$$\eta = \text{logit}(p) \stackrel{\text{def}}{=} \log\frac{p}{1-P} \Rightarrow p = \sigma(\eta)$$

$$\eta = \text{probit}(p) \stackrel{\text{def}}{=} \Phi^{-1}(p) \Rightarrow p = \Phi(\eta)$$

where $p \stackrel{\text{def}}{=} \Pr(R = 1|X = x)$, $\eta = b + w'x$, $\sigma(\eta) = \frac{1}{1+\exp(-\eta)}$ is the sigmoid function, and $\Phi(x) = P(Z \leq x)$, $Z \sim N(0, 1)$, is the cdf of the standard Normal. The logit and probit distributions are very similar (see Figure 2), and differ only in the tails; essentially, the cumulative normal dies off as $e^{-x^2}$, whereas the sigmoid dies off more slowly as $e^{-x}$.

Although probit has a nice interpretation as a noisy threshold unit ($R = 1$ iff $y > Z$), the logistic distribution has several advantages:

- It can be well-motivated from a statistical viewpoint [Jor95].
- There is an efficient method for fitting its parameters, called the Iterative Reweighted Least Squares (IRLS) algorithm [MN83, JJ94b] (a form of Newton-Raphson).
- There is a good approximation method for converting it to potential form (see Section 6).
- It generalizes to multi-valued discrete variables as follows:

$$\Pr(R = i|X = x) = \frac{\exp(w_i'x + b_i)}{\sum_j \exp(w_j'x + b_j)}$$

This is called the softmax (multinomial logit) function. Note that softmax for binary variables is equivalent to the logistic function when $w = w_1 - w_0$ and $b = b_1 - b_0$, since $\Pr(R = 1|X = x) = \frac{e^{w_1'x+b_1}}{e^{w_0'x+b_0}+e^{w_1'x+b_1}} = \frac{1}{1+e^{(w_0-w_1)'x+(b_0-b_1)}}$.

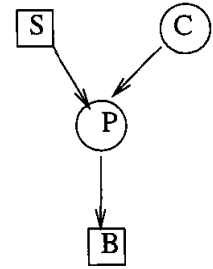

Figure 1: The crop network. Circles represent continuous (scalar) nodes, squares represent discrete (binary) nodes. This example is from [BKRK97].

In the softmax function, $w_i$ is the normal vector to the $i$'th decision boundary, and $b_i$ is its offset. The magnitude of $w_i$ determines the steepness of the curve: a large magnitude corresponds to a hard threshold (steep curve), and a small magnitude corresponds to a soft threshold. In the limit as $|w_i| \to \infty$, the sigmoid approaches a step function; in the limit as $|w_i| \to 0$, the sigmoid approaches a uniform distribution.

It turns out that linear Gaussians and softmax are both special cases of Generalized Linear Models (GLIMs): see [MN83] or [JJ94b] for details. Although we can use GLIMs as CPDs for observed nodes (see Section 8), in general it is difficult to use them for hidden nodes, at least if we restrict ourselves to exact inference.

### 2.1 Example

As a simple example of some of the distributions we have described, consider the network in Figure 1. In this model, the price (P) of a certain crop, say wheat, is assumed to decrease linearly with the amount of crop (C) produced that year, on the assumption that a glut reduces prices. But if the government artificially subsidises prices ($S = 1$), the price will be raised by a fixed amount. In addition, the consumer is likely to buy ($B = 1$) if the price drops below 5 units (see Figure 2). This model will be used for the experiments in Section 10 with the parameter values shown below.

| Node | Distribution | Params. |
|---|---|---|
| S | CPT | $p = 0.3$ |
| C | Gaussian | $\mu = 5, \Sigma = 1$ |
| P | CG | $\mu_0 = 10, \mu_1 = 20,$ |
| | | $W_{0,1} = -1, \Sigma_{0,1} = 1$ |
| B | Logistic | $w = -1, b = 5$ |

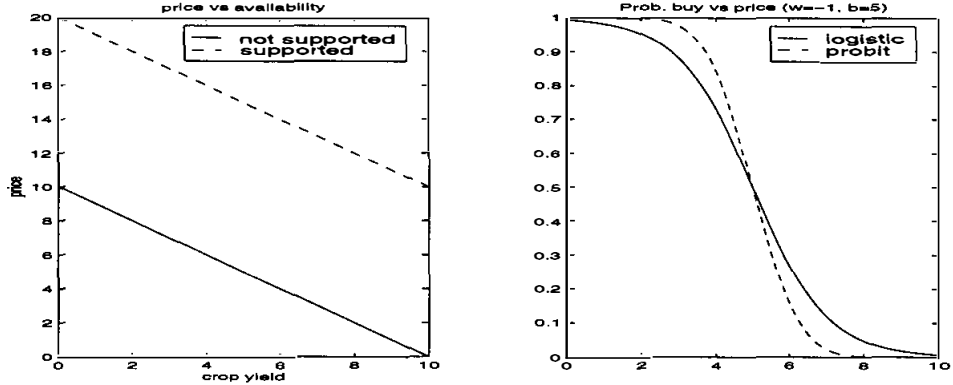

Figure 2: Left: The expected price decreases linearly with the crop yield, $E[P|C, S = 0] = 10 - C$, and is shifted up by a constant if the price is artificially supported, $E[P|C, S = 1] = 20 - C$. Right: The probability someone will buy the crop decreases as the price rises above the threshold of 5. We plot $\sigma(wx + b)$ and $\Phi(wx + b)$, where $x$ is the price, $w = -1$ and $b = 5$.

## 3 The junction tree algorithm

In this section, we give a brief overview of the junction tree algorithm (see e.g., [HD94] for details), before discussing the aspects of it which are specific to hybrid networks. This summary is meant to provide a road map for the rest of the paper.

In the junction tree algorithm, we first perform the following graph-theoretic steps in order.

- Moralize the original graph $G$, i.e., connect together all parents who share a common child, and then drop the directionality of the arcs. This will result in an undirected graph, $G_M$.
- Choose an elimination ordering $\pi$, e.g., according to the heuristics discussed in [Kja90].
- Let all nodes be initially unmarked. For each node in order $\pi$, mark it and join all its unmarked neighbors. This will result in a triangulated graph, $G_T$. (See [BG96] for more effective ways to triangulate a graph.)
- Find the maximal cliques in $G_T$; call them $\mathcal{C}$.
- Build an undirected weighted graph $G_J$ whose nodes are the cliques $\mathcal{C}$ and where the weight of the edge from clique $i$ to clique $j$ is $|C_i \cap C_j|$. Let $T$ be a maximal spanning tree of $G_J$ [JJ94a].
- Add a separator node $S$ to each edge $(i, j)$ of $T$ such that $S = C_i \cap C_j$.
- Pick an arbitrary node in $T$ as root.

In Section 5, we discuss the changes that need to be made to the above steps in the case of hybrid networks.

After building the junction tree "shell", we perform the following numerical steps in order. These steps involve the potentials associated with each clique and separator; how to represent and operate on such potentials is discussed in Section 4.

- For each clique and separator in $T$, initialize its potential to the identity element.
- For each node $X$ in $G$, find a clique $C$ in $T$ that contains $X$ and its parents, convert $X$'s CPD to a potential (see Sections 6 and 7), and multiply it onto $C$'s potential.
- Optionally, we can now perform a global propagation, to convert the potentials into joint form; these can then be saved for later reuse, so that we can avoid repeating this initialization step. (In the approach to evidence that we discuss in Section 8, it is not possible to do a propagation before the evidence has arrived.)
- For each node $X$ for which we have evidence, find a clique $C$ that contains $X$, and multiply in the evidence (see Section 8).
- For each clique $V$ in postorder (i.e., children before parents), make $W$ absorb from $V$, where $W$ is $V$'s parent in $T$. (This is called the "collect evidence" phase.) $W$ absorbs from $V$ via separator $S$ by performing the following operations:
  - $\phi^*(S) = \sum_{V \setminus S} \phi(V)$.
  - $\phi^*(W) = \phi(W) \times (\phi^*(S) \div \phi(S))$.

  where $\phi$ is a potential, the $*$ superscript denotes the new or updated potential, $\sum$ represents the marginalization operator, $\times$ the multiplication operator, and $\div$ the division operator. (We say that $V$ sends a "message" to $W$.)
- For each clique $V$ in preorder (i.e., parents before children), make $W$ absorb from $V$, for each child $W$ of $V$. (This is called the "distribute evidence" phase.)

## 4 Hybrid clique potentials

When all the variables in a clique are discrete, we can represent its potential using a table (multidimensional array); when all the variables are Gaussian, we can represent the potential as a quadratic form; and when some of the variables are discrete, and some are Gaussian, we can use a table of quadratic forms. We now explain the quadratic form representation; see [LW89, Lau92, Ole93, Lau96] for details.

A Gaussian clique potential can either be represented in familiar moment form

$$P(x; p, \mu, \Sigma) = p \exp\left[-\tfrac{1}{2}(x - \mu)'\Sigma^{-1}(x - \mu)\right]$$

or the more convenient canonical form

$$P(x; g, h, K) = \exp\left[g + x'h - \tfrac{1}{2}x'Kx\right]$$

We can convert from canonical to moment form (provided $K$ is full rank) as follows:

$$\begin{aligned}
\Sigma &= K^{-1} \\
\mu &= \Sigma h \\
\log p &= g - \tfrac{1}{2}\log|K| + \frac{n}{2}\log(2\pi) + \tfrac{1}{2}\mu' K \mu
\end{aligned}$$

We can always convert from moment to canonical form.

A CG potential is just a list of such Gaussian potentials, one for each value of the discrete variables. Note that, by using a logarithmic representation of the constant factor, we are assuming the $p(i)$ is never non-zero. To get around this, we need to additionally store an indicator variable, $\chi(i)$, which is 1 iff this discrete value has positive support. (One advantage of the logarithmic representation is that it is unlikely to underflow even if we have a lot of evidence.)

We now define how to peform the fundamental operations of extension, multiplication/division, and marginalization on CG potentials.

Extension is the operation of ensuring that two potentials are defined on the same set of variables. For the continuous variables, we must make sure the size of each vector and matrix is the same, by inserting 0s where necessary.[1] For the discrete variables, we must make sure both potentials have the same number of table entries, duplicating where necessary.

We can now define multiplication of two CG potentials, $\phi_1(W)$ and $\phi_2(V)$, as follows.

- Convert both potentials to canonical form, if necessary.
- Extend them to the same domain, if necessary.
- Compute the following for each discrete entry:

$$(g_1, h_1, K_1) \times (g_2, h_2, K_2) = (g_1+g_2, h_1+h_2, K_1+K_2)$$

Division is similar, except we use $-$ instead of $+$.

Marginalization is harder. Let us first consider the case of pure Gaussian potentials. Suppose we want to compute $\phi(x_2) = \int_{x_1} \phi(x_1, x_2)$. We first convert $\phi$ to canonical form (if necessary)[2], and then partition it into the components being kept and the components being marginalized over:

$$h = \begin{pmatrix} h_1 \\ h_2 \end{pmatrix}, \quad K = \begin{pmatrix} K_{11} & K_{12} \\ K_{21} & K_{22} \end{pmatrix}$$

The new canonical characteristics are as follows:

$$\begin{aligned}
\hat{g} &= g + \tfrac{1}{2}(p\log(2\pi) - \log|K_{11}| + h_1' K_{11}^{-1} h_1) \\
\hat{h} &= h_2 - K_{21} K_{11}^{-1} h_1 \\
\hat{K} &= K_{22} - K_{21} K_{11}^{-1} K_{12}
\end{aligned}$$

Now let us turn to the CG case. We first marginalize over the continuous variables, and then the discrete ones.

---

[1] We assume there is a canonical ordering for the entries within each vector/matrix/table.

[2] It is much easier to marginalize in moment form (just extract the relevant components of $\mu$ and $\Sigma$); however, it is not always possible to convert to moment form.

However, this does not necessarily reduce the size (number of table entries) of the CG potential. For example, consider the potential $\phi(x, y, i, j)$ where $x$ and $y$ are continuous scalar variables, and $i$ and $j$ are discrete binary variables; hence $\phi$ is a mixture of four (two dimensional) Gaussians. If we marginalize over $y$ and $j$, the result will be $\phi(x, i) = \sum_j \int_y \phi(x, y, i, j)$ which is still a mixture of four (one dimensional) Gaussians: marginalization has not made the potential any smaller (in terms of the number of discrete components). Now suppose we multiply this potential by $\psi(x, z, i, k)$, where $z$ is a scalar and $k$ is a binary variable — the result will now be a mixture of 8 (two dimensional) Gaussians, instead of just 4, since for each value of $i$ and $k$, $\phi$ contains two Gaussians. Hence, as we propagate messages, the potentials become mixtures with more and more components.

To avoid the exponential blow-up in the size of the potentials, we adopt the standard approximation of "collapsing" a mixture of Gaussians to a single Gaussian, using the following formulas (this is called "weak marginalization"):

$$\begin{aligned}
p(i) &= \sum_j p(i, j) \\
\mu(i) &= \sum_j p(j|i) \mu(i, j) \\
\Sigma(i) &= \sum_j p(j|i) \big[\Sigma(i, j) + \\
&\qquad (\mu(i, j) - \mu(i))(\mu(i, j) - \mu(i))'\big]
\end{aligned}$$

where $p(j|i) = p(i, j) / \sum_j p(i, j)$. (In our example, we collapse a mixture of 2 Gaussians to a single Gaussian for each value of $i$.) This is the best CG approximation (in the sense of minimizing KL divergence) to the true marginal (see e.g., [Lau96, p. 162] for a proof). In particular, it gives the correct first and second order moments, i.e., $E[x|i]$ and $\text{Var}[x|i]$ will be the same for the weak marginal and the true marginal.

Note that if the parameters of the Gaussian are independent of the discrete variable being marginalized over (i.e., $\mu(i, j) = \mu(i)$ and $\Sigma(i, j) = \Sigma(i)$) — for example, because the discrete variable is not a parent or child of the Gaussian variables but just happens to "live" in the same clique — then this process is exact, and is called "strong marginalization".

## 5 Junction trees with strong roots

The non-closure of CG-potentials under marginalization of discrete variables means that we have to be careful how we construct the junction tree. In particular, we need to be able to convert to moment form before we perform any discrete marginalizations. The relevant theory is discussed in [Lau92, Lau96]; here, we just summarize the main results.

We define a strong root as any node $R$ (in the junction tree) which satisfies the following property: for any pair $V, W$ of neighbors on the tree with $W$ closer to $R$ than $V$, we have

$$(V \setminus W) \subseteq \Gamma \lor (V \cap W) \subseteq \Delta$$

where $\Gamma$ are all the continuous variables and $\Delta$ are all the discrete variables. In other words, when a separator be-tween two neighboring cliques is not purely discrete, all the variables in the clique furthest away from the root which are not in the separator are continuous. If a graph is triangulated and does not have any paths between two discrete vertices passing through only continuous vertices (i.e., a "forbidden path" of the form $D - C - D$), then there is always at least one strong root [Lei89]; such graphs are called decomposable, marked graphs (marked just means there are two types of nodes).

For example, consider Figure 1. Moralization adds an arc between $S$ and $C$; the resulting graph is then already triangulated, and has cliques $SCP$ and $PB$, so the junction tree is $SCP - P - PB$. Note that this has a forbidden path from $S$ to $B$, and hence there is no strong root. However, if we add an extra arc between $S$ and $B$ after the moralization step, to eliminate the forbidden path, the junction tree becomes $SCP - SP - SPB$. Here, $SPB$ is a strong root, since $V \setminus W = \{S, C, P\} \setminus \{S, P, B\} = \{C\} \subseteq \Gamma$.

A sufficient condition to ensure there is a strong root is to eliminate all the continuous nodes before the discrete ones when triangulating. For example, if we use the elimination order $\pi = (C, P, S, B)$, we get the strong junction tree above.

We need a junction tree with a strong root is to ensure that when we send messages up to the root (during the collect evidence phase), all the marginalizations will be strong, so that when we subsequently send messages back from the root (during the distribute evidence phase), neighboring potentials will be consistent.

The reason the first pass results in strong marginals is easy to see: for any pair $V, W$ of neighbors on the tree with $W$ closer to the strong root than $V$, when we compute $\phi^*(S) = \sum_{V \setminus S} \phi(V)$, we are only performing integrations, since the only variables in $V$ which are not in $W$ are continuous (by definition). When we have to marginalize out a discrete variable, say $I$, we can always integrate out any variables which depend on it, say $X$, first (i.e., we can compute $\sum_i \int_x \phi(x, i)$ instead of $\int_x \sum_i \phi(x, i)$), and hence avoid the need to collapse the mixture of Gaussians.

The reason the second pass results in consistent potentials is also easy to see. Suppose that $W$ absorbed from $V$ on the first pass, so $\phi(S) = \sum_{V \setminus W} \phi(V)$. On the backwards pass, we compute $\phi^*(V) = \phi(V) \times (\phi^*(S) \div \phi(S))$, so

$$\sum_{V \setminus W} \phi^*(V) = (\phi^*(S) \div \phi(S)) \times \sum_{V \setminus W} \phi(V)$$

$$= \phi^*(S) = \sum_{W \setminus V} \phi^*(W)$$

(Note that we are justified in pulling the ratio outside the sum only because the marginalization over $V \setminus W$ is strong.)

The disadvantage of requiring a strong root is that, in general, adding extra links to remove forbidden paths will increase the size of the cliques, as we saw above. One can always choose to ignore the strong root requirement, although this risks incurring additional inaccuracies of unknown magnitude. Fortunately, as we will see in Section 9, the *effective* size of a clique is determined only by the number of *hidden* nodes it contains, so adding extra links to observed nodes does not increase the computational complexity.

## 6 Converting CPDs to potentials

We now discuss how to convert CPDs into potentials for Gaussian nodes with Gaussian and/or discrete parents, and for discrete nodes with discrete and/or Gaussian parents.

For a Gaussian node with Gaussian parents, we can create a canonical potential as follows.

$$\begin{aligned} P(y|x) &= C(\Sigma) \exp\left[-\tfrac{1}{2}(y - \mu - Wx)'\Sigma^{-1}(y - \mu - Wx)\right] \\ &= \exp\left[-\tfrac{1}{2}\begin{pmatrix} x & y \end{pmatrix}\begin{pmatrix} W\Sigma^{-1}W' & -W\Sigma^{-1} \\ -\Sigma^{-1}W' & \Sigma^{-1} \end{pmatrix}\begin{pmatrix} x \\ y \end{pmatrix}\right. \\ &\quad \left. + \begin{pmatrix} x & y \end{pmatrix}\begin{pmatrix} -W'\Sigma^{-1}\mu \\ \Sigma^{-1}\mu \end{pmatrix} - \tfrac{1}{2}\mu'\Sigma^{-1}\mu + \log C(\Sigma)\right] \end{aligned}$$

Hence we set the canonical characteristics to

$$\begin{aligned} g &= -\tfrac{1}{2}\mu'\Sigma^{-1}\mu + \log C(\Sigma) \\ h &= \begin{pmatrix} -W'\Sigma^{-1}\mu \\ \Sigma^{-1}\mu \end{pmatrix} \\ K &= \begin{pmatrix} W\Sigma^{-1}W' & -W\Sigma^{-1} \\ -\Sigma^{-1}W' & \Sigma^{-1} \end{pmatrix} \end{aligned}$$

This generalizes the result in [Lau92] to the case of vector-valued nodes. In the scalar case, $\Sigma^{-1} = 1/\sigma$, $W = w$, and $n = 1$, so the above becomes

$$\begin{aligned} g &= \frac{-\mu^2}{2\sigma^2} - \tfrac{1}{2}\log(2\pi\sigma^2) \\ h &= \frac{\mu}{\sigma^2}\begin{pmatrix} -w \\ 1 \end{pmatrix} \\ K &= \frac{1}{\sigma^2}\begin{pmatrix} ww' & -w \\ -w' & 1 \end{pmatrix} \end{aligned}$$

From this, we see that $K$ is rank 1; hence we may not be able to represent the *initial* potential on a clique in moment form (although after we have propagated evidence, each potential represents a *joint* probability density, which can always be converted to moment form).

For a Gaussian node with discrete parents, we get a $(g, h, K)$ triple of the above form for each value of the discrete parents.

For discrete nodes with discrete parents, we can convert any CPD into a CPT (i.e., we can compute $\Pr(R = i | Q = j)$, even if the distribution is specified implicitly), and hence can convert it to a tabular potential. Of course, such a transformation might lose some local conditional independence information, which might have been exploited to speed up inference. For some kinds of CPDs, such as noisy causal independence models [RD98], there are ways to expose the local structure graphically, which makes it easier to exploit in the junction tree framework, but we don't discuss this issue here.

Finally, we discuss the case of discrete nodes with Gaussian parents in the next section.

## 7 The variational approximation

We can convert the logistic function to a canonical Gaussian potential by using the following variational lower bound

462 Murphy

[JJ96] (see Appendix A for the derivation):

$$\Pr(R = r | X = x) = \sigma(w'x + b)$$
$$\geq \sigma(\xi) \exp\left[(A - \xi)/2 + \lambda(\xi)(A^2 - \xi^2)\right]$$

where $A = (2r - 1)(w'x + b)$, $\lambda(\xi) = (\frac{1}{2} - \sigma(\xi))/2\xi$, and $r \in \{0, 1\}$. Notice that this is quadratic in $x$, and hence we can represent it as a canonical potential:

$$g = \log \sigma(\xi) + \frac{1}{2}(2r - 1)b - \frac{1}{2}\xi + \lambda(\xi)(b^2 - \xi^2)$$
$$h = \frac{1}{2}(2r - 1)w + \lambda(\xi)2bw$$
$$K = -2\lambda(\xi)ww'$$

We call this representation VG, for Variational Gaussian. If the discrete node also has discrete parents, we get a $(w_i, b_i)$ pair for each discrete parent value, and the resulting potential will be a mixture of VGs (MVG).

The advantage of the variational approximation is that it allows us to represent the potential as a Gaussian, and hence perform marginalization in closed form. The need to do this arises even in sparsely connected models (ones which have small clique size). This is in contrast to the more common use of variational methods, which is to approximate inference in models which are too dense to solve exactly (see [JGJS98] for a review).

With any approximation method, it is natural to ask how good the approximation is. Although a quadratic function is a poor approximation to a sigmoid, the *joint* probability $P(X, R)$ (where $X$ is Gaussian and $R$ is logistic) is well-approximated by a Gaussian (see Figure 3). In fact, the approximation is *exact* when $\xi = (2r - 1)(w'x + b)$.

If $X$ is hidden, the optimal value of $\xi$ cannot be computed. However, we can guess an initial value, and then iteratively adjust it to increase the quality of the approximation. As in EM [NH98], at each iteration we set $\xi$ to the value that maximizes the expected complete-data log-likelihood, where the expectation is computed using the parameter values of the previous iteration. This results in the following update (see Appendix B for the derivation):

$$\xi^2 = E\left[(w'x + b)^2\right] = w'(\Sigma + \mu\mu')w + 2bw'\mu + b^2$$

where the posterior distribution on $X$ is $X \sim N(\mu, \Sigma)$. (The update equation does not specify whether to take the positive or negative square root. However, this ambiguity turns out not to matter, since $\Pr(R|X;\xi)$ is symmetric in $\xi$.)

Choosing a good initial estimate of $\xi$ is important (see Section 10). The procedure we use is as follows. We walk down the graph and compute the mean and variance of each node (if it is continuous), or its most probable value (if it is discrete), based only on the evidence and assignments above it. Then, when we get to a logistic node, we can look up $\mu$ and $\Sigma$ of its parents, and plug them into the equation above for $\xi$.

For example, consider the crop network and suppose only $B$ is observed. We set $S = 0$ (since the $S$ node is more likely to be off than on), $E[C] = 5$, and $E[P] = E[P|S = 0] = 10 - 5 = 5$, i.e., we use the mixture component corresponding to the most probable value of $S$. (This turned out to be better than collapsing the mixture of Gaussians at $P$, using the distribution over $S$ for the weights.)

We can also derive the *upper* bound $P(r|x) \leq \exp(\alpha A - H_2(\alpha))$, where $\alpha$ is another variational parameter and $H_2(p) \stackrel{\text{def}}{=} -p \log p - (1 - p) \log(1 - p)$ is the binary entropy function. We use the lower bound because (1) it is tighter (since it is a second-order approximation), and (2) for learning, we want to maximize a lower bound on the likelihood [NH98]. However, the upper bound can be used in conjunction with the lower bound to filter out runs of MCMC which result in marginals which fall outside the bounds, as in [JJ99].

Note that we can also exploit the quadratic approximation to fit the parameters of the logistic node, $w$ and $b$, using linear regression, instead of the slower IRLS (Iteratively Reweighted Least Squares) procedure, as noted in [Tip98].

Finding a good variational approximation for the softmax distribution is a problem we are currently working on. In this paper, we only consider the logistic distribution (i.e., binary nodes). However, we can always use $k$ binary nodes to encode (in a distributed fashion) the value of a single node $R$ with $2^k$ possible values (see e.g., [Tip98]).

## 8 A new approach to handling evidence

The "traditional" approach to handling evidence in the junction tree framework is as follows (see e.g., [HD94]). Let us start by considering the case where all the potentials are discrete. First we create a junction tree with the potentials initialized to 1s, then we multiply on all the CPDs. When evidence arrives, e.g., we observe that $Q = i$, we find any clique that contains $Q$ and multiply it by a potential of the form $(0, \ldots, 1, \ldots, 0)$, where the 1 is in the $i$'th position.[3] This sets to 0 any entries incompatible with the evidence. Finally, we do a propagation to restore global consistency. Now each clique potential contains the joint probability of its variables and the evidence $e$, e.g., $\phi(Q = i, R = j) = \Pr(Q = i, R = j, e)$. This can be normalized to obtain the likelihood of the evidence, $\Pr(e) = \sum_{i,j} \phi(i, j)$, and the posterior, $\Pr(Q = i, R = j|e) = \phi(i, j)/\Pr(e)$.

When we have Gaussian potentials, we initialize to 0s, and follow a similar procedure, except now we must multiply *every* potential (including separators) that contains the nodes for which we have evidence, since the dimensionality of the vectors and matrices will be reduced. For example, suppose we observe $Y = y$; then $\phi(X, Y)$ becomes

$$\phi^*(x) = \exp\left[g + (x' \ y')\begin{pmatrix} h_X \\ h_Y \end{pmatrix} - \frac{1}{2}(x \ y)\begin{pmatrix} K_{XX} & K_{XY} \\ K_{YX} & K_{YY} \end{pmatrix}\begin{pmatrix} x \\ y \end{pmatrix}\right]$$
$$= \exp\left[(g + h_Y'y - \frac{1}{2}y'K_{YY}y) + x'(h_X - K_{XY}y) - \frac{1}{2}x'K_{XX}x\right]$$

This generalizes the equation in [Lau92] to the case of vector-valued nodes.

There are several problems with the traditional approach:

- For discrete variables with many possible values (e.g., HMMs with large codebooks), we may create huge initial clique potentials, only to subsequently set most of the entries to zero. The technique of evidence shrinkage [HD94] and zero compression [JA90] can help reduce the inefficiency of manipulating such sparse potentials, but it would be better not to create them in the first place.

---
[3]This is sometimes called "hard" evidence. "Soft" or "virtual" evidence would consist of a distribution over $Q$'s possible values.

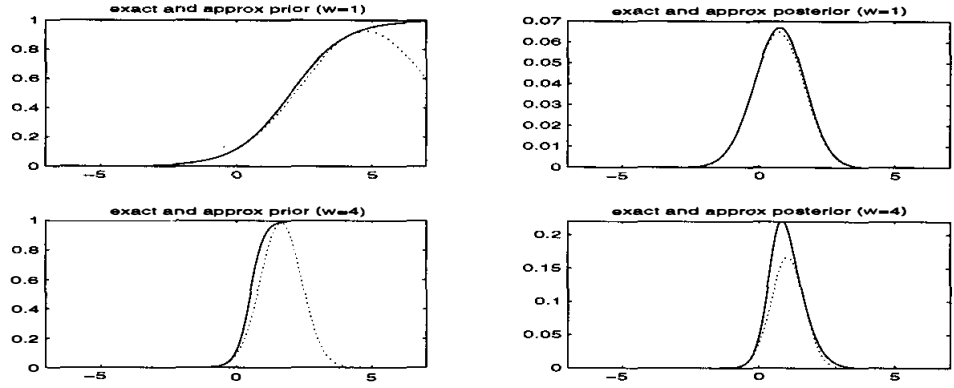

Figure 3: The variational approximation gets poorer as the logistic function becomes steeper (more deterministic). On the left we plot $\Pr(R = 1 | x)$ for the exact (solid) and approximate (dotted) logistic function, using $b = -2$ and $w = 1$ (top) or $w = 4$ (bottom), and the optimal $\xi$ value. On the right, we plot $\Pr(R = 1 | x) \Pr(x)$, where $\Pr(x) = N(x; 0, 1)$.

- We need to have a way of converting the CPD of each node into potential form. This makes it impossible to use many kinds of distributions. Also, we might want to create a conditional model $\Pr(Y|X)$, and not associate any parameters with $X$ since it is always an observed input (as in linear regression). This is not possible with the traditional approach.

- There is an annoying asymmetry in the handling of observations on discrete and continuous nodes. For the former, we only need to modify one potential, but for the latter, we must modify all potentials that contain the observed nodes. In addition, it is difficult to do the book-keeping when we change the size of each Gaussian potential.

There is a very simple solution to all these problems: create the initial clique potentials *after* the evidence has arrived! Then the potentials only have to be defined on the hidden nodes: the observed nodes just contribute a constant factor to the value of clique potential, and don't take up any space.

For example, consider a softmax node with a parent $X$ whose value is observed to be $x^*$. We can convert this to a CPT and thence into a discrete potential by computing $\Pr(R = i | X = x^*) = \text{softmax}(x^*, i)$ for each $i$. Similarly, consider an HMM with Gaussian output. We can create the evidence-specific observation matrices by computing $\Pr(Y_t = y_t^* | Q = i) = N(y_t^*; \mu_i, \Sigma_i)$ for each hidden state $i$ and each time step $t$. This is the sense in which we can use arbitrary conditional densities on observed nodes.

The type of potential that we need to use in the junction tree is determined by the type of hidden nodes that are left. If all the hidden nodes in a clique are discrete (D), we can represent its potential with a table; if they are all Gaussian (G), we can use a Gaussian; otherwise, we must use a Mixture of Gaussians (MG). If there is one potential of type D and another of type G, all the potentials will be converted to type MG for compatibility (i.e., so they can absorb from one another). Similarly, if one is of type MG and another is of type D or G, the latter will be converted to MG. (That is, all the cliques are "raised" to their least common ancestor in the type hierarchy, which has MG above both D and G.)

The disadvantage of the new approach to handling evidence is that it is not incremental, i.e., when new evidence arrives, we cannot just update a small part of the junction tree, but instead must combine the new and old evidence, and rerun the whole inference algorithm. In addition, the new approach cannot handle retraction of evidence or soft evidence. On the other hand, it is simple to combine the new technique with the old, so that the "core" findings can be handled in the new way, and nodes for which we have soft evidence, or which we might want to just temporarily instantiate, can be handled in the old way.

## 9 Computational complexity of inference

It is possible to marginalize and multiply/divide a Gaussian potential in $O(n^3)$ time, where $n$ is the size of the potential (i.e., the number of scalar variables in the clique), whereas these operations on a discrete potential take time linear in the number of entries in the table, which is *exponential* in the number of discrete variables in the clique. Hence large cliques only impose a high computational cost if they contain many *discrete* variables. This is the reason why people have been able to exactly solve large linear Gaussian models, such as Kalman filters, without having to resort to the kinds of approximations that are used in the discrete Bayes net community.

By using the new approach to handling evidence, we only need to worry about cliques that contain many *hidden* discrete variables. More precisely, if we partition the nodes into a hidden and observed set, $V = H \cup O$, or into a discrete and continuous set, $V = D \cup C$, then the cost of inference in a hybrid network is

$$O\left(\sum_{c \in \mathcal{C}} \left[\left(\prod_{x \in c \cap H \cap D} |x|\right) \times \left(\sum_{x \in c \cap H \cap C} |x|\right)^3\right]\right)$$

where $\mathcal{C}$ is the set of cliques, $|x|$ is the number of values node $x$ can take on (if it is discrete) or its length (if it is a continuous-valued vector). (See [MA98] for a more detailed discussion of the complexity of the junction tree algorithm for discrete networks.)

## 10 Experimental results

To see how accurate the variational approximation is, we compared the junction tree algorithm (as implemented in BNT[4]) to Gibbs Sampling (as implemented in BUGS[5]) on the network shown in Figure 1. We generated 20 random examples from the joint distribution encoded by the network (using the exact setting of $\xi$), and computed the posterior distributions over the hidden variables, for each possible pattern of evidence, using the junction tree or BUGS.

For BUGS, we used a "burn-in" of 2000 iterations, and then sampled for 10,000 iterations. (Similar results were achieved using a burn-in of just 1000 plus 1000 iterations, and also using 1000 samples from likelihood weighting [SP90].) For the junction tree, we updated the variational parameters until the relative change in log-likelihood dropped below 0.001; when $S$ was observed, so $P$ had a unimodal distribution, this took 2–3 iterations; when $S$ was hidden, so $P$ had a bimodal distribution, this took 7–9 iterations. BUGS (implemented in compiled Modula 2) took 17 and 12 minutes of real time, for 10,000 and 1000 iterations respectively, and the junction tree (implemented in interpreted Matlab) took 2.5 minutes. (Times are for a Sun Ultra Sparc 2.)

The results are shown in Table 2. $\Delta(S)$ is $|E_j[S] - E_b[S]|$ averaged over 20 trials, and similarly for $\Delta(C)$, $\Delta(P)$, and $\Delta(B)$, where $E_j$ and $E_b$ are the expected values (conditioned on the evidence) computed using the junction tree and BUGS respectively. (Standard deviation is in brackets.) A dash means the variable was observed, so inference was not necessary. (Note that, since $S$ is a binary random variable, $E[S] = \Pr(S = 1)$ and similarly for $B$.)

When $P$ is observed (with value, say, $p$), we can perform inference exactly (because $P(P = p|S, C)$ can be represented as a CG distribution), and so the non-zero values of $\Delta$ are due to finite sampling effects in BUGS (which can always be driven to 0 by taking more samples). When $P$ is hidden, we need to use the variational approximation, and so the non-zero values of $\Delta$ are partly due to errors incurred by this approximation, and partly due to finite sampling effects.

The results indicate that the variational approximation does well except in cases where both $S$ and $B$ are hidden (rows 14 and 16). This is because, in this case, the posterior on $P$ is bimodal: there will be a peak near $P = 5$ and one near $P = 15$, corresponding to $S = 0$ and $S = 1$ respectively. Furthermore, since $B$ is not observed, it is hard tell which one is more likely (apart from the prior, of course). Note that observing $C$ is not particularly helpful, since it is d-separated from $S$, and in any case rarely deviates from its mean (since $\text{Cov}[C] = 1$). When $P$ is bimodal, it is difficult to choose a good initial estimate of $\xi$, which can cause the algorithm to converge to a poor local maximum (of the lower-bound on the log-likelihood). When we "cheated" by starting the variational algorithm off with the correct value of $\xi$ (i.e., $\xi = (2r-1)(w'x+b)$, usingthe true values of $P = x$ and $R = r$), we found, not surprisingly, that the variational method did very well.[6]

---
[4]Bayes Net Toolbox. See www.cs.berkeley.edu/ murphyk/Bayes/bnt.html.
[5]Bayesian inference Using Gibbs Sampling. See www.mrc-bsu.cam.ac.uk/bugs/.

## 11 Future work

We are currently working on ways to improve the quality of the approximation in the case that the distribution over the parents of the logistic node is multimodal. We also want to extend the variational approximation to softmax nodes. In the future, we intend to apply these techniques to hybrid Dynamic Bayesian Networks, which can be thought of as an extension to the traditional Switching Kalman Filter model [BSL93]; in particular, the methods in this paper allow the mode switches to be determined by the hidden continuous state, instead of occuring "spontaneously".

## 12 Acknowledgments

I would like to thank Mike Jordan for useful discussions. This work was supported by grant number ONR N00014-97-1-0941.

## A Derivation of the quadratic lower bound to the logistic function

In this section, we derive a quadratic lower bound on the sigmoid function
$$\sigma(x) = (1 + e^{-x})^{-1}$$
For details, see [Jaa97].

Consider first
$$1 + e^x = e^{x/2}(e^{-x/2} + e^{x/2}) = e^{x/2 + \log(e^{-x/2} + e^{x/2})}$$
$$\stackrel{\text{def}}{=} e^{x/2 + \hat{f}(x)}$$
where $\hat{f}(x) = \log(e^{-x/2} + e^{x/2})$ is symmetric, and a concave function of $x^2$.

Now, for any concave function $f(x)$, it is easy to see that
$$f(x) \leq \left[(x - \xi)\frac{\partial}{\partial \xi}f(\xi) + f(\xi)\right]$$
$$\stackrel{\text{def}}{=} \bar{\lambda}(\xi)x - \bar{\lambda}(\xi)\xi + f(\xi)$$
where $\bar{\lambda}(\xi) = \frac{\partial}{\partial \xi}f(\xi)$, i.e., any tangent line to the function is an upper bound, which is tight when $\xi = x$.

Let $f(x^2) \stackrel{\text{def}}{=} \hat{f}(x)$, and let $\xi^2$ be the variational parameter indicating the location of the tangent, so that
$$\hat{f}(x) = f(x^2) \leq \bar{\lambda}(\xi)x^2 + f(\xi^2) - \xi^2\bar{\lambda}(\xi)$$

---
[6]In fact, it did better than any exact method could, since by using the optimal $\xi$, we "leaked" information about the true values of $P$ and $R$. For example, in case 16, when there are no observations, the posteriors should be equal to the priors (i.e., $E[S] = 0.3$, $E[C] = 5$, $E[P] = 0.7(10-5) + 0.3(20-5) = 8$, $E[B] = 0.35$), and yet the cheating method computed different estimates of these quantities for each example, indicating that it had knowledge of the 'true values'; furthermore, these estimates were closer to the true values than the exact priors.

|    | S | C | P | B | $\Delta(S)$ | $\Delta(C)$ | $\Delta(P)$ | $\Delta(B)$ |
|----|---|---|---|---|-------------|-------------|-------------|-------------|
| 1  | o | o | o | o | -           | -           | -           | -           |
| 2  | h | o | o | o | 0.0000 (0.0000) | -       | -           | -           |
| 3  | o | h | o | o | -           | 0.0033 (0.0002) | -       | -           |
| 4  | h | h | o | o | 0.0000 (0.0000) | 0.0034 (0.0003) | - | -       |
| 5  | o | o | h | o | -           | -           | 0.0152 (0.0101) | -       |
| 6  | h | o | h | o | 0.0000 (0.0000) | -       | 0.0063 (0.0037) | -       |
| 7  | o | h | h | o | -           | 0.0110 (0.0062) | 0.0176 (0.0137) | - |
| 8  | h | h | h | o | 0.0000 (0.0000) | 0.0352 (0.0145) | 0.0424 (0.0218) | - |
| 9  | o | o | o | h | -           | -           | -           | 0.0018 (0.0021) |
| 10 | h | o | o | h | 0.0000 (0.0000) | -       | -           | 0.0026 (0.0030) |
| 11 | o | h | o | h | -           | 0.0022 (0.0003) | -       | 0.0019 (0.0017) |
| 12 | h | h | o | h | 0.0000 (0.0000) | 0.0006 (0.0003) | - | 0.0023 (0.0022) |
| 13 | o | o | h | h | -           | -           | 0.2286 (0.1455) | 0.2800 (0.1862) |
| 14 | h | o | h | h | 0.2957 (0.0000) | -       | 2.8897 (0.3186) | 0.3745 (0.1788) |
| 15 | o | h | h | h | -           | 0.2756 (0.1530) | 0.5506 (0.3033) | 0.3812 (0.2258) |
| 16 | h | h | h | h | 0.3015 (0.0000) | 0.3337 (0.0000) | 2.3247 (0.0000) | 0.3480 (0.0000) |

Table 2: Experimental results for the crop network using the junction tree. **Cols. 1–4**: 'o' means a variable is observed, 'h' means it is hidden. **Cols. 5–8**. $\Delta(S)$ is $|E_j[S] - E_b[S]|$ averaged over 20 trials, and similarly for $\Delta(C)$, $\Delta(P)$, and $\Delta(B)$, where $E_j$ and $E_b$ are the expected values (conditioned on the evidence) computed using the jtree and BUGS respectively. (Standard deviation in brackets.) A dash means the variable was observed, so inference was not necessary. See text for details.

where

$$\tilde{\lambda}(\xi) = \frac{\partial}{\partial \xi^2} f(\xi^2) = \frac{\partial}{\partial \xi^2} \hat{f}(\xi)$$

Using the substitution $u = \xi^2$ we find

$$\tilde{\lambda}(\xi) = \frac{1}{4\xi} \tanh(\xi/2) = (\sigma(\xi) - \tfrac{1}{2})/2\xi$$

Now,

$$\log \sigma(-X) = -\log(1 + e^X) = -(X/2 + \hat{f}(X))$$

so we get the following *lower* bound on the logistic function:

$$\begin{aligned}
\log \sigma(X) &= X/2 - \hat{f}(X) \\
&\geq X/2 - \tilde{\lambda}(\xi)X^2 - f(\xi^2) + \xi^2 \tilde{\lambda}(\xi) \\
&\geq X/2 + \lambda(\xi)X^2 + \log \sigma(\xi) - \xi/2 - \xi^2 \lambda(\xi)
\end{aligned}$$

where $\lambda(\xi) = -\tilde{\lambda}(\xi)$. Using the fact that $\Pr(R = 0|X = x) = 1 - \Pr(R = 1|X = x) = \sigma(-(w'x + b))$, we get the final result.

## B  Derivation of update formula for the variational parameter

To find the optimal value of $\xi$, we iteratively maximize a lower bound on the expected complete-data log-likelihood, as in EM [NH98]. The only term which depends on $\xi$ is $E[\log P(R = r|X = x; \xi)]$, where the expectation is w.r.t. all the observed data and the $\xi$ from the previous iteration. Differentiating, we get

$$\frac{\partial}{\partial \xi} E[\log P(R = r|X = x)]$$
$$= \frac{\partial \lambda(\xi)}{\partial \xi} \frac{\partial}{\partial \lambda} E\left[\log \sigma(\xi) + (A - \xi)/2 + \lambda(\xi)(A^2 - \xi^2)\right]$$
$$= \frac{\partial \lambda(\xi)}{\partial \xi} (E[A^2] - \xi^2)$$

Since $\lambda(\xi)$ is monotonically increasing in $|\xi|$, the maximum is obtained at

$$\xi^2 = E[A^2] = E\left[E[A^2|R = r]\right]$$

$$\begin{aligned}
&= P(R = 0)E[(-1)^2(w'x + b)^2] \\
&\quad + P(R = 1)E[(1)^2(w'x + b)^2] \\
&= E[(w'x + b)'(w'x + b)] \\
&= E[\mathrm{tr}(x'ww'x)] + 2bw'E[x] + b^2 \\
&= \mathrm{tr}(w'E[xx']w) + 2bw'E[x] + b^2 \\
&= w'(\mathrm{Cov}[X] + E[X]E[X'])w + 2bw'E[x] + b^2 \\
&= w'(\Sigma + \mu\mu')'w + 2bw'\mu + b^2
\end{aligned}$$

where $E$ and $P$ are taken w.r.t. all the data and the previous $\xi$.

Note that this derivation is slightly more general than the one in [JJ96], since we allow $R$ to be hidden; however, the net result turns out to be the same.

## References

[AA96]    S. Alag and A. Agogino. Inference using message propogation and topology transformation in vector Gaussian continuous networks. In *Proc. of the Conf. on Uncertainty in AI*, 1996.

[BG96]    A. Becker and D. Geiger. A sufficiently fast algorithm for finding close to optimal junction trees. In *Proc. of the Conf. on Uncertainty in AI*, 1996.

[BKRK97] J. Binder, D. Koller, S. J. Russell, and K. Kanazawa. Adaptive probabilistic networks with hidden variables. *Machine Learning*, 29:213–244, 1997.

[BSL93]   Y. Bar-Shalom and X. Li. *Estimation and Tracking: Principles, Techniques and Software*. Artech House, 1993.

[CF91]    K. C. Chang and R. M. Fung. Symbolic probabalistic inference with continuous variables. In *Proc. of the Conf. on Uncertainty in AI*, pages 77–85, 1991.

[CF95]    K. C. Chang and R. M. Fung. Symbolic probabilistic inference with both discrete and continuous variables. *IEEE Trans. on Systems, Man, and Cybernetics*, 25(6):910–917, 1995.

[Dec98]   R. Dechter. Bucket elimination: a unifying framework for probabilistic inference. In M. Jordan, editor, *Learning in Graphical Models*. MIT Press, 1998.

[DM95] E. Driver and D. Morrel. Implementation of continuous Bayesian networks usings sums of weighted Gaussians. In *Proc. of the Conf. on Uncertainty in AI*, pages 134–140, 1995.

[FG96] N. Friedman and M. Goldszmidt. Discretizing continuous attributes while learning bayesian networks. In *Machine Learning: Proceedings of the International Conference*, 1996.

[GRS96] W. Gilks, S. Richardson, and D. Spiegelhalter. *Markov Chain Monte Carlso in Practice*. Chapman and Hall, 1996.

[HD94] C. Huang and A. Darwiche. Inference in belief networks: A procedural guide. *Intl. J. Approx. Reasoning*, 11, 1994.

[JA90] F. Jensen and S. K. Andersen. Approximations in Bayesian belief universes for knowledge-based systems. In *Proc. of the Conf. on Uncertainty in AI*, 1990.

[Jaa97] T. Jaakkola. *Variational Methods for Inference and Estimation in Graphical Models*. PhD thesis, MIT, 1997.

[JGJS98] M. I. Jordan, Z. Ghahramani, T. S. Jaakkola, and L. K. Saul. An introduction to variational methods for graphical models. In M. Jordan, editor, *Learning in Graphical Models*. MIT Press, 1998.

[JJ94a] F. V. Jensen and F. Jensen. Optimal junction trees. In *Proc. of the Conf. on Uncertainty in AI*, 1994.

[JJ94b] M. I. Jordan and R. A. Jacobs. Hierarchical mixtures of experts and the EM algorithm. *Neural Computation*, 6:181–214, 1994.

[JJ96] T. Jaakkola and M. Jordan. A variational approach to Bayesian logistic regression problems and their extensions. In *AI + Statistics*, 1996.

[JJ99] T.S. Jaakkola and M.I. Jordan. Variational probabilistic inference and the QMR-DT network. *J. of AI Research*, 10, 1999.

[Jor95] M. I. Jordan. Why the logistic function? A tutorial discussion on probabilities and neural networks. Technical Report 9503, MIT Computational Cognitive Science Report, August 1995.

[Kja90] U. Kjaerulff. Triangulation of graphs – algorithms giving small total state space. Technical Report R-90-09, Dept. of Math. and Comp. Sci., Aalborg Univ., Denmark, 1990. Available at www.cs.auc.dk/ uk.

[KK97] A. V. Kozlov and D. Koller. Nonuniform dynamic discretization in hybrid networks. In *Proc. of the Conf. on Uncertainty in AI*, 1997.

[Lau92] S. L. Lauritzen. Propagation of probabilities, means and variances in mixed graphical association models. *J. of the Am. Stat. Assoc.*, 87(420):1098–1108, December 1992.

[Lau96] S. Lauritzen. *Graphical Models*. OUP, 1996.

[Lei89] H.-G. Leimer. Triangulated graphs with marked vertices. *Annals of Discrete Mathematics*, 41:311–324, 1989.

[LW89] S. L. Lauritzen and N. Wermuth. Graphical models for associations between variables, some of which are qualitative and some quantitative. *Annals of Statistics*, 17:31–57, 1989.

[MA98] R. J. McEliece and S. M. Aji. The generalized distributive law. *IEEE Trans. Info. Theory*, 1998. submitted.

[MN83] McCullagh and Nelder. *Generalized Linear Models*. Chapman and Hall, 1983.

[NH98] R. M. Neal and G. E. Hinton. A new view of the EM algorithm that justifies incremental and other variants. In M. Jordan, editor, *Learning in Graphical Models*. MIT Press, 1998.

[Ole93] K. G. Olesen. Causal probabilistic networks with both discrete and continuous variables. *IEEE Transactions on Pattern Analysis and Machine Intelligence*, 3(15), 1993.

[RD98] I. Rish and R. Dechter. On the impact of causal independence. Technical report, Information and Computer Science, UCI, 1998.

[SAS94] R. Shachter, S. Andersen, and P. Szolovits. Global conditioning for probabilistic inference in belief networks. In *Proc. of the Conf. on Uncertainty in AI*, 1994.

[SK89] R. Shachter and C. R. Kenley. Gaussian influence diagrams. *Managment Science*, 35(5):527–550, 1989.

[SP90] R. D. Shachter and M. A. Peot. Simulation approaches to general probabilistic inference on belief networks. In *Proc. of the Conf. on Uncertainty in AI*, volume 5, 1990.

[Tip98] M. Tipping. Probabilistic visualization of high-dimensional binary data. In *Neural Info. Proc. Systems*, 1998.